# Segmentation of histological images and fibrosis identification with a convolutional neural network

Xiaohang Fu[1] · Tong Liu[2] · Zhaohan Xiong[1] · Bruce H. Smaill[1] · Martin K. Stiles[3] · Jichao Zhao[1]

**Abstract** Segmentation of histological images is one of the most crucial tasks for many biomedical analyses including quantification of certain tissue type. However, challenges are posed by high variability and complexity of structural features in such images, in addition to imaging artifacts. Further, the conventional approach of manual thresholding is labor-intensive, and highly sensitive to inter- and intra-image intensity variations. An accurate and robust automated segmentation method is of high interest. We propose and evaluate an elegant convolutional neural network (CNN) designed for segmentation of histological images, particularly those with Masson's trichrome stain. The network comprises of 11 successive convolutional – rectified linear unit – batch normalization layers, and outperformed state-of-the-art CNNs on a dataset of cardiac histological images (labeling fibrosis, myocytes, and background) with a Dice similarity coefficient of 0.947. With 100 times fewer (only 300 thousand) trainable parameters, our CNN is less susceptible to overfitting, and is efficient. Additionally, it retains image resolution from input to output, captures fine-grained details, and can be trained end-to-end smoothly. To the best of our knowledge, this is the first deep CNN tailored for the problem of concern, and may be extended to solve similar segmentation tasks to facilitate investigations into pathology and clinical treatment.

Xiaohang Fu
xfu697@aucklanduni.ac.nz

Jichao Zhao
j.zhao@auckland.ac.nz

[1] Auckland Bioengineering Institute, The University of Auckland, Auckland 1142, New Zealand

[2] Department of Cardiology, Second Hospital of Tianjin Medical University, and Tianjin Key Laboratory of Ionic-Molecular Function of Cardiovascular Disease, Tianjin Institute of Cardiology, Tianjin 300201, P.R. China

[3] Waikato Hospital, Hamilton 3204, New Zealand

## 1 Introduction

Accurate segmentation of biomedical images is fundamental for quantitative analysis. However, this task is challenging due to characteristically high inhomogeneity and complexity of features in such images, as well as inter- and intra-plane artifacts introduced as a result of the imaging procedure and methodology, such as lighting. Manual thresholding is the most prevalent method for segmenting biomedical images, for example to identify fibrosis or scarring in histology [1, 2]. Whilst straightforward in principle, this approach is labor-intensive, time-consuming, and may involve tedious re-adjustments of thresholds [3, 4]. Thresholds are also highly sensitive to subject-dependent biases, as well as inter- and intra-image intensity variations (since spatial information is not accounted for) [5, 6]. Thus, a variety of thresholds is commonly necessary for one image set. For such reasons, manual thresholding may not be feasible for large datasets, especially those containing considerable variability in intensity, contrast, or brightness.

Interpretation of raw pixel intensities to image meaning or context is no trivial task for algorithms. A slight difference in image features such as illumination may be negligible to humans, but can result in a disparate algorithmic outcome. Numerous methods have been established to separate an image into groups displaying similar features, and thereby identify the class object of each pixel. Earlier segmentation techniques rely on distinguishing edges, regions, or textures [6]. However, for image data with highly irregular structural features, heterogeneous illumination, or variable coloring of similar objects, considerable pre- or postprocessing is required, thus rendering such techniques unattractive and largely unsuitable.

In recent years, machine learning for computer vision has advanced extensively, emerging as a powerful tool for a wide range of image recognition problems [7–10]. Machine learning methods can be generally classed as unsupervised or supervised. In the former, the algorithm identifies patterns in the input without learning from example data annotated with desired outputs (ground truth). Contrastingly, supervised models are trained on labelled data, learning rules to produce outputs from inputs. Unsupervised methods including $k$-means clustering [11, 12], mean-shift clustering [13], and Markov random fields [14], as well as earlier supervised approaches such as support vector machines [10] have previously been employed to segment histological images. However, these methods typically suffer the requirement of supplementary algorithms (e.g. for postprocessing [12]) to complete the segmentation objective, or additional domain expertise to define and extract suitable features from images, which are often based on strong assumptions about the data.

Convolutional neural networks (CNNs) are generating great enthusiasm particularly in computer vision. Conceptualized in the 1980s [15], CNNs were biologically inspired by the visual cortex; neurons fire in response to certain features or patterns in their local receptive fields, thereby acting as spatial filters [16]. CNNs effectively map highly complex relationships between the input and desired output (such as those of shapes and colors present in images), through interconnected stacks of nonlinear functions (most fundamentally convolutions). Contrary to alternative supervised learning approaches such as support vector machines, there is no manual hand-crafting or fine-tuning of useful features in the input. CNNs can achieve impressive performance and directly handle complex data with minimal manual effort. A CNN-based approach is fully automated and trained models are reusable after establishment.

Although the inception of neural networks was a few decades ago, deep networks with multiple stacked layers are a relatively recent development; brought about through progress in parallelized computation using GPUs, solutions to hindrances associated with training deep neural networks (such as rectified linear units (ReLU) for the vanishing gradient problem [17]), and the availability of very large datasets. Deep CNNs have proved to be powerful tools in a wide array of image-related applications, excelling in image classification [7, 18, 19], handwriting recognition [20], object localization [21], and scene understanding [22, 23]. This technique has also successfully extended to semantic pixel-wise labeling and the biomedical domain in tasks such as image segmentation [24–26], detection [27], cell tracking [8], and computer-aided diagnosis [9].

Robust and automated segmentation methods that can overcome the inherent challenges of biomedical image segmentation are of great demand, especially for applications conventionally relying on a manual approach. An example is fibrosis identification in histology, a critical task in many key fields of clinical research including kidney failure [28], lung injury [29], hepatitis B [30], sinoatrial node [1], and atrial fibrillation [31].

Atrial fibrillation is the most common type of cardiac arrhythmia, associated with significant healthcare costs, reduced quality of life, morbidity and mortality [32]. The basic mechanisms behind its initiation and maintenance remain elusive, but accumulating recent evidence indicate that diabetes mellitus (DM) is a strong independent risk factor [31–34], and that atrial fibrosis or scarring (characterized by excessive extracellular matrix proteins including collagen) induced under diabetic conditions contributes considerably to arrhythmogenicity [31, 35, 36]. Quantification and comparison of atrial fibrotic remodeling under DM against controls will assist in illuminating the precise mechanisms underlying DM-induced atrial fibrillation. This requires segmentation of fibrosis from myocytes and background in a cardiac histological section, differentiated via the well-accepted Masson's trichrome stain (Fig. 1).

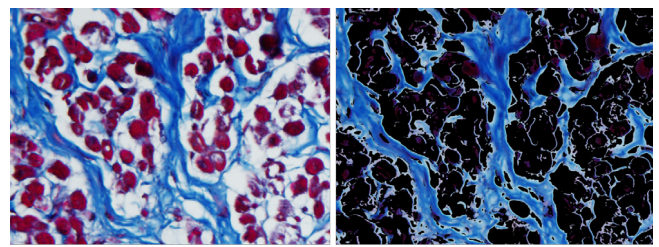

**Fig. 1** Representative segmentation of fibrosis. *Left* typical original RGB image of left atrial tissue from a diabetes mellitus (DM) rabbit model imaged with Masson's trichrome stain at 40× magnification (image size 0.33 mm × 0.25 mm, pixel spatial resolution 161.25 nm × 161.25 nm); red, white and blue respectively indicate healthy myocytes, fat or extracellular space, and fibrosis. *Right* segmented fibrotic regions via a thresholding approach shown in blue

In this paper, we propose a CNN for automated segmentation of stained histology images into a required number of tissue types, with particular focus on quantifying fibrosis in cardiac sections. To the best of our knowledge, this is the first CNN designed for this application. The deep CNN displays state-of-the-art segmentation accuracy with drastically fewer parameters, and substantially greater efficiency. More importantly, our proposed CNN architecture can be extended to other similar segmentation tasks to facilitate understanding of certain diseases and to aid targeted clinical treatment. We also make our source code freely available online for the benefit of potential users.

## 2 Methods and experiments

### 2.1 Overview of CNNs for segmentation

During a forward pass through a CNN, characteristics specific to certain structures in an input image (such as intensity and spatial information) are discerned by trainable filters in convolutional layers. Convolutional filters (typically $3 \times 3$ pixels) sweep across the entire visual field of the input volume by a constant stride, thus allowing the CNN to detect features in the input regardless of their exact position. Convolutional layers compute dot products between learnable filter weights and a corresponding region of the input slice, generating activation or feature maps. Thus, activations contain contextual representations about the image, with larger values indicating better resemblance between weights and intensity patterns in the receptive field of a filter. Stacking of many convolutional layers can amplify the capacity of a CNN to identify features of greater complexity from a larger field of view in the input. This ability may be further enhanced by the inclusion of other nonlinear operations such as ReLU. Typically, pooling layers are inserted to subsample outputs from convolutions, and feature maps progressively increase in abstraction through the network as low-level information is compounded over many subsampling operations. Weights are automatically adapted during supervised learning to optimize sensitivity to certain features of relevance and maximize accuracy of predictions to ground truth. A loss function quantifying the disparity between predictions and ground truth is minimized via gradient descent, and errors are backpropagated through the network to modify weights accordingly.

In contrast to CNNs for image classification where the output is a single class (for example "dog"), CNNs for image segmentation require dense per-pixel classification and spatial localization of classes in the form of an output segmentation map. The recent advancement of deep CNNs for segmentation was pioneered by the development of a fully convolutional network (FCN) by Long *et al.* [37]. Currently, the most prevalent and successful CNNs for segmentation are inspired by the scheme of FCN, adapting configurations originally designed for classification to perform per-pixel labeling by substituting fully connected layers with convolutions [8, 23, 38, 39]. Such architectures constitute of downsampling and upsampling stages, also known respectively as an encoder and decoder. The image is first downsampled by a series of convolutions and max pooling to obtain low resolution feature maps, which are then upsampled by deconvolution (also named convolutional transpose or fractionally strided convolution), generating the localization of classes desired. In some cases, concatenation between intermediate feature maps, or unpooling [40] are incorporated to improve final resolution. However, additional upsampling layers introduce more trainable parameters in such architectures, which are commonly on the order of tens to hundreds of millions [8, 23, 37, 38]. Thus, they are susceptible to overfitting particularly when available data is scarce, and may be difficult to train end-to-end. Additionally, results may be too coarse due to subsampling by multiple max pooling operations. These configurations may not be optimal for all types of images and segmentation goals due to unique feature properties.

### 2.2 Proposed 11-layer CNN

The proposed CNN architecture is illustrated in Fig. 2. The input image to the CNN may be of any resolution and number of channels, and does not diminish in resolution through the network. The output is a volume of the same depth as the number of desired segmentation classes (3 for our problem i.e. fibrosis, myocyte, and background), in which per-pixel probabilities for each class are stored. The CNN is suitable for any number of classes. The final segmentation map constituting of colors which correspond to the predicted class for each pixel can be obtained by determining the argmax-index along the depth axis of the output volume.

Our CNN consists of 11 convolutional layers. The first 9 involve $3 \times 3$ filters with 64 channels, succeeded by 2 layers of $1 \times 1$ convolutions with 3 channels. All convolutions have a stride of 1. We incorporate further nonlinearity by appending ReLU activation after all convolutional layers in the network, performing $\max(0, x)$ element-wise to $x$ (the output from a convolution), thereby converting all negative values to 0. ReLUs are a non-saturating and computationally simple method to capture more complex features of the input data, and has been shown to greatly accelerate CNN learning during stochastic gradient descent [7, 17].

We normalize input distributions of every convolutional layer to zero mean and unit variance,

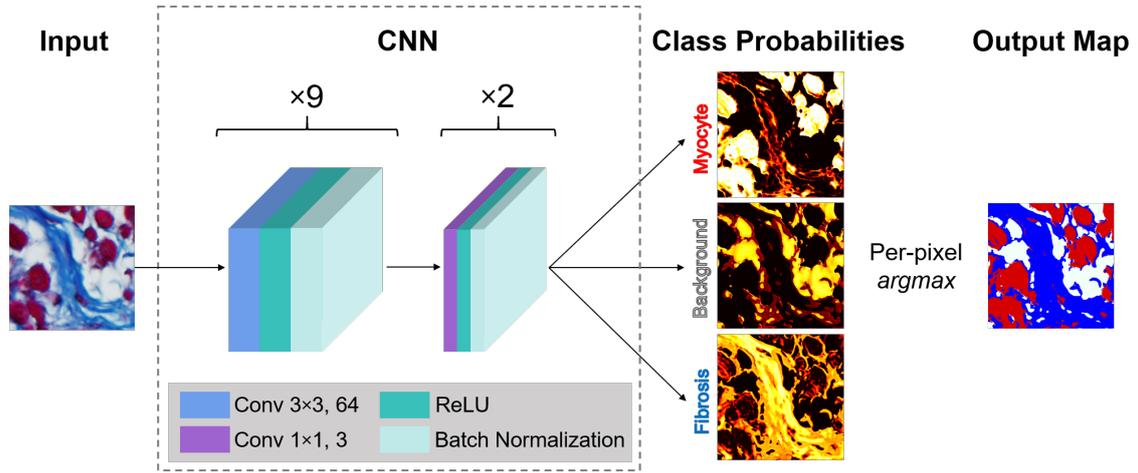

**Fig. 2** Architecture of the proposed 11-layer convolutional neural network (CNN) for dense per-pixel classification. It consists of 9 layers of 3 × 3 convolutions with 64 channels, followed by 2 layers of 1 × 1 convolutions with 3 channels. All convolutions are performed with stride 1, and are appended by rectified linear unit (ReLU) activation, then batch normalization. The output of the CNN contains per-pixel probabilities for each class of the same resolution as the input image

addressing the problem of internal covariate shift with batch normalization [41]:

$$BN(x_i) = \gamma \left( \frac{x_i - \mu_B}{\sqrt{\sigma_B^2 + \varepsilon}} \right) + \beta \quad (1)$$

where $BN$ is the batch normalizing transform, $x_i$ is an activation from mini-batch $B$, $\mu_B$ is the mini-batch mean, $\sigma_B^2$ is the mini-batch variance, $\gamma$ and $\beta$ are the learnable scale and shift terms respectively, and $\varepsilon$ is a small constant to prevent division by zero. As a result, we observe greatly accelerated model learning and improved evaluation performance. Batch normalization is positioned after ReLU nonlinearity, as our experiments indicated that this provides a larger performance increase over the inverse arrangement. We do not add biases to the outcomes of convolutions, as they would effectively be negated by the mean subtraction step in batch normalization. Furthermore, we do not incorporate dropout [42] as model regularization is provided by batch normalization, following the practice in [41].

Disparate to the typical encoder-decoder state-of-the-art deep architectures for segmentation, we do not downsample the image and reduce lateral resolution (e.g. via pooling), then upsample to produce a dense output. Although subsampling may improve the robustness of a model to small shift variants of the input, spatial localization suffers since representations are coarser. As we observe high variance of low-level features within and between local regions in histology images of our kind, we exclude pooling to prevent smoothing of fine-grained features, and consequently gain several other advantages.

With the use of zero padding to preserve X-Y spatial size of intermediate volumes after the otherwise contracting convolutions, lateral resolution is maintained throughout the CNN from input to output. Hence our segmentation network is free of upsampling layers (e.g. deconvolution). This substantially reduces the number of trainable parameters in the network and its susceptibility to overfitting.

Additionally, the number of feature maps in deep CNNs typically double after halving lateral dimensions by stride 2 max pooling, following the highly influential VGG-type designs [18]. Contrastingly, volume depth in our CNN is maintained at 64 until the final 2 layers, which contain 3 feature channels each. This configuration provides favorable performance whilst compensating for the omitted benefits to computational efficiency by subsampling.

Overall, we can build a more elegant and efficient architecture containing fewer parameters by deviating from convention; eliminating pooling and upsampling, and not increasing the number of feature channels through the network. Our proposed architecture contains 300 thousand trainable parameters, which is 447, 103, and 840 times less than FCN (~134 million), U-Net (~31 million), and DeconvNet (~252 million), respectively [8, 37, 38]. The size of our trained TensorFlow model on disk is just 3.5 MB.

Through experimentation, we found stacking two 1 × 1 convolution – ReLU – batch normalization blocks at the end of the CNN an inexpensive way to provide additional nonlinearity to better capture relevant features in the receptive field and slightly boost performance, over one single such block. Depth of our CNN architecture may be further increased by inserting convolutional layers with 64 feature maps of the same fashion. Our experiments indicate that this does not drastically improve segmentation performance, however each additional layer introduces 36,864 trainable parameters. As we prefer efficiency and fewer parameters over a slight benefit to accuracy, we selected a total of 11 convolutional layers in our network. We also experimented with two other modifications to our network, in the form of dilated convolutions [43] with successive dilation factors of 1-1-2-4 in four early layers, or residual-style connections [19] through the CNN. Results indicate that these variants are about equal in performance.

## 2.3 Experiments

Supervised learning of a CNN is typically based on a large dataset consisting of images and corresponding labels. Training is followed by an evaluation phase where predictions are produced from the trained model, and performance is analyzed. The following subsections present the detailed steps we implemented to address the image segmentation problem for fibrosis identification, and to benchmark segmentation performance of the proposed approach.

### 2.3.1 Dataset

The dataset consists of 72 images, 36 each from control and DM groups, all 2064 × 1536 pixels (width × height) in size with 3 color channels (RGB). The histological images are courtesy of Fu *et al.*, methods detailed in [31]. Briefly, left atrial sections of control and DM (induced by alloxan monohydrate) Japanese rabbits were stained with Masson's trichrome and imaged with an Olympus DP72 at 40× objective magnification. Each pixel has a spatial resolution of 161.25 nm × 161.25 nm.

### 2.3.2 Manual thresholding and ground truth

Manual thresholding is presently the conventional approach for identifying tissue type and fibrosis in histology sections with Masson's trichrome stain. In these images, fibrosis is colored blue and myocytes red, while white is regarded as background. We first meticulously applied manual thresholding to the original images by utilizing manually toned regional thresholds and detailed touch-ups, generating segmentation ground truth. To ensure our ground truth is precise, we validated the segmentation results with experts in the field [31].

To facilitate segmentation, the images were preprocessed prior to thresholding via histogram normalization, to standardize the minimum and maximum intensities in each RGB channel to 0 and 255, respectively. We employed multiband thresholding for higher segmentation accuracy, involving combinations of different thresholds across channels to isolate classes, whilst ensuring no pixel was unclassed or multi-classed. Thresholds were selected through repeated empirical trials and visual validation.

### 2.3.3 Data augmentation

We randomly selected 24 original images (12 each from control and DM groups) to construct the training set. Since performance of neural networks is generally improved with more training data, the amount of available data for training was amplified by augmentation, a technique popular in the field of deep learning for image classification [7, 44]. Augmentation also benefits to reduce overfitting, improving model invariance to adjustments negligible to classification outcome. We applied the following independent transformations identically to each original image and its corresponding labels, then randomly sampled 48 × 48 patches from augmented forms (number of patches in brackets):
- Rotation by 90º (450), 180º (900), or 270º (450);
- Flipping along the horizontal (450) or vertical axis (450);
- Sinusoidal warping (900);
- Shearing affine transformation (900).

A maximum of 900 patches were obtained from one augmented version, which is roughly two-thirds the image area in pixel count.

To ensure that each patch captured adequate information and avoided extremely biased proportions of pixels for any classes, we excluded from the training set the highest 4/9 of patches as ranked by the standard deviation of their class proportions. We then randomly discarded 96 patches such that the total number of patches is divisible by 128, the size of the mini-batches during gradient descent. The training set consisted of 59,904 48 × 48 patches in total, about 138 million pixels. Prior to training, we randomized the order of training data with the intent of achieving smoother convergence during gradient descent.

Due to random sampling, proportions of each class in the training set roughly reflected those in the test set, with myocytes in the greatest prevalence (44%), followed by background (32%), and fibrosis the lowest (24%). Data class imbalance is a major problem in supervised classifiers, detrimentally impacting minority classes in particular [45, 46]. A vast array of strategies has been devised for overcoming this recurrent problem (although with variable success), including oversampling, undersampling, retaining natural proportions in learning examples, data synthesis, and class-weighted loss functions [45]. With the intent of improving segmentation performance for fibrosis, we carried out another training of the proposed CNN with a different training set consisting of approximately balanced class proportions (35% myocyte, 34% background, and 31% fibrosis), obtained by oversampling fibrosis. This training set consisted of the same number of 48 × 48 patches as the standard training set.

We also assessed the capability of our approach to withstand variations in color and brightness typical of Masson's trichrome-stained histology sections, by evaluating performance of the proposed CNN on color-adjusted test images. The maximum extents of the alterations are visualized in Fig. 3. Further, we performed an independent training of the CNN using a color augmented version of the standard training set. We randomly selected 40% of patches to undergo contrast adjustment via the red channel, and a different 40% via the blue channel. Training images were color-adjusted by a random degree up to the predefined maximum limits.

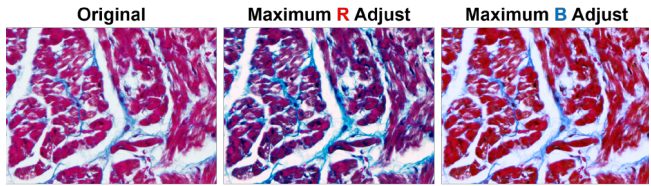

**Fig. 3** Maximum extents of color augmentation applied via the red (*center*) and blue (*right*) channels

### 2.3.4 CNN training

Training is an iterative process in which (i) training data are fed into the model in batches, (ii) predictions are produced by the current model in a forward pass, (iii) errors between predictions and ground truth labels are computed, (iv) errors are backpropagated through the network, (v) parameter corrections for all neurons are computed, and (vi) parameters are updated to minimize errors. A single cycle over the entire training set is an epoch, and typically multiple epochs are needed for convergence.

We utilize the TensorFlow framework [47] to implement the CNN, and perform end-to-end training from scratch. For the first iteration of training, we initialize weights per recommendations by He *et al.* [48]. We sample weights from a normal distribution centered on zero, truncated at two standard deviations from the mean, with variance $2/n$, where $n$ is the product of the number of input feature channels, filter height, and filter width for a given layer.

For preprocessing during model training and evaluation, we subtract training set RGB means from the corresponding channel, then normalize each channel to unit variance via division of its standard deviation.

During learning, we compute cross-entropy as the loss function, which measures dissimilarity between per-pixel class distributions of ground truth and estimated probabilities. First, we normalize predicted probabilities to unit sum for each pixel via the softmax function in Eq. 2, where $z$ contains predicted probabilities for $K$ classes, and $f_j$ corresponds to the $j$-th element in the vector of softmax probabilities $f$.

$$f_j(z) = \frac{e^{z_j}}{\sum_{k=1}^{K} e^{z_k}} \qquad (2)$$

We then determine loss of each pixel point $L_i$ by Eq. 3, where $y_i$ is the correct class, and $j$ is the index along vector $f$. We compute loss as the mean loss across all pixels in a mini-batch.

$$L_i = -\log\left(\frac{e^{f_{y_i}}}{\sum_{j=1}^{J} e^{f_j}}\right) \qquad (3)$$

Training is carried out via gradient descent with training images in mini-batches set at size 128. Weights are updated every mini-batch, and one revised model is produced every epoch. We minimize cross-entropy loss during stochastic optimization via Adam [49], which utilizes adaptive learning rates for smooth convergence. We set the three parameters of Adam, the upper bound learning rate, and exponential decay rates for the first and second moment estimates, to their default values of 0.001, 0.9, and 0.999, respectively. Order of mini-batches is re-randomized every epoch except the last batch, to simplify tracking for visualization of training progress. We end training when mean Dice similarity coefficient (DSC) for the test set does not increase by at least 1% after 20 further epochs from its current best epoch. The best model is the one producing the highest test mean DSC.

### 2.3.5 Evaluation

Our primary segmentation performance metric is the widely-adopted DSC, which assesses spatial overlap by combining precision and recall in the form of a harmonic mean [50]. We also measure intersection over union (IoU), a common metric for semantic segmentation, and the evaluation standard in Pascal VOC2012 [51]. Both are measures of overlapping areas of mutual class assignment, but differ slightly in formulation. DSC and IoU scores are within (0, 1), with higher values indicating better performance.

Preliminary for computation of the two metrics is the construction of a confusion matrix between ground truth and predictions, allowing the tallying of true positive (TP), false positive (FP), false negative (FN), and true negative (TN) outcomes predicted by the classifier for each class. We compute DSC and IoU scores for class *c* by Eqs. 4 and 5, respectively:

$$DSC_c = \frac{2 \cdot TP_c}{2 \cdot TP_c + FP_c + FN_c} \quad (4)$$

$$IoU_c = \frac{TP_c}{TP_c + FP_c + FN_c} \quad (5)$$

Class scores are subsequently averaged for each image, yielding image DSC and image IoU. We distill overall performance of a classifier into a single value for each metric by averaging respective image scores across the 48-image test set, yielding mean DSC and mean IoU. We also report mean scores for each class determined in an analogous manner. To select the top-performing model from each training instance, each model produced predictions for the entire test set, and the basis for model selection was its overall mean DSC. We visually scrutinized all segmentation outputs from machine learning approaches to confirm their accuracy.

**2.3.6 Comparison with previous methods**

We compare performance of the proposed architecture for the per-pixel classification task at hand with two well-adopted CNNs for segmentation, FCN-8 [37] and U-Net [8]. FCN-8 is the most refined version of the FCNs, and a landmark development in recent progress of CNNs for image segmentation, achieving a 20% improvement in performance against traditional approaches on standard datasets. U-Net was designed specifically for biomedical image segmentation, and has proved its superiority by winning several contests. The network can achieve high performance with very few training data and has become widely popular, adapted for many applications [52–54].

For learning and testing, we used identical data as our proposed architecture, trained all networks from scratch as outlined in Section 2.3.4, and followed the same evaluation procedures in Section 2.3.5. The only modification we made to the original U-Net was the use of zero padding during convolutions, to preserve the lateral size of input volume at such steps. For FCN-8 and U-Net, we initialized all biases to zero. We employed dropout during training for the two fully connected layers in FCN-8, and all convolutional layers except the last in U-Net. We experimentally determined dropout rates of 0.5 and 0 (no dropout) respectively for FCN-8 and U-Net to yield the best performance on the test set. We present their performance scores accordingly.

We also compare segmentation performance against that of *k*-means clustering, a widely-utilized unsupervised machine learning algorithm which partitions *N* unlabeled observations (pixels in the case of images) into *K* groups [55]. In the case of RGB image segmentation into $K = 3$ classes, three centroids exist, each located at RGB intensities deemed optimal by the algorithm. We also performed *k*-means in Lab space after conversion from RGB. The performed iterative process is as follows:

i. Random initialization of cluster centroids ($C_k$, $k = 1…K$);
ii. Assignment of data points ($x_n$, $n = 1…N$) to clusters with current centroids of minimum Euclidean distance $\|x_n - C_k\|$ away;
iii. Computation of new centroids using updated groupings;
iv. Repetition of steps ii and iii until convergence.

To avoid results in local minima, the algorithm was performed three times, and the final partitions with the lowest sum of distances from all points to centroids was selected. Since *k*-means is unsupervised, the classes of segmentations are unknown. Our strategy is to produce all six permutations possible with three classes and select the one which scores highest in overall test mean DSC.

## 3 Results and discussion

### 3.1 Proposed CNN

In Fig. 4, we present the curves of training accuracy and loss, and test accuracy over epochs. The training and test accuracy curves converge approaching epoch number 40, when training met our criterion for termination.

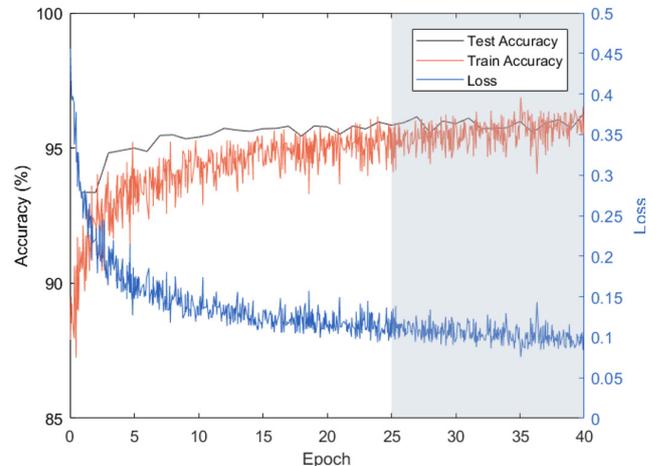

**Fig. 4** Test accuracy, training accuracy, and loss over training epochs of the proposed CNN. Grey highlights convergence of test accuracy

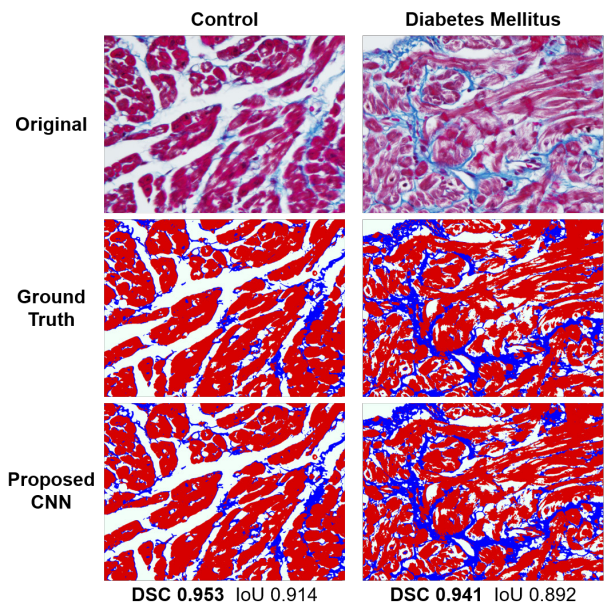

**Fig. 5** Representative segmentations by the proposed CNN for control and DM images. Mean Dice similarity coefficient (DSC) and mean intersection over union (IoU) are indicated for each corresponding image group

The proposed system achieved high in segmentation performance, scoring a test mean DSC of 0.947. Expectedly, predictions are visually almost identical to ground truth. Performance on images of the control group was slightly above the DM group (Fig. 5).

We observed imbalanced class performance for all the neural networks we tested (including other architectures); e.g., 0.980 vs. 0.907 in DSC for myocytes and fibrosis (Table 1). Identification of fibrosis was weaker relative to myocyte and less so to background, which reflects the relative prevalence of each class in the data. This problem was not alleviated by the technique of oversampling minority class (fibrosis) examples by training with roughly balanced class proportions. In fact, evaluation performance of the proposed CNN reduced slightly (0.944 mean DSC), with background and fibrosis suffering relatively more (decreased by 0.004) than the dominant myocytes (decreased by 0.002). Thus, our results suggest higher per-class and overall segmentation performance when training and test set class proportions are more aligned. Hence, it may be beneficial to be aware of potentially disparate distributions between training and testing data, which could demand re-training or fine-tuning of the model.

By testing the trained network with color-adjusted image sets (Fig. 3), we show the ability of the proposed method to tolerate different image color characteristics (such as brightness and contrast), and be invariant to the exact colors of structures (Fig. 6). For the test images which were altered by modifying blue channel

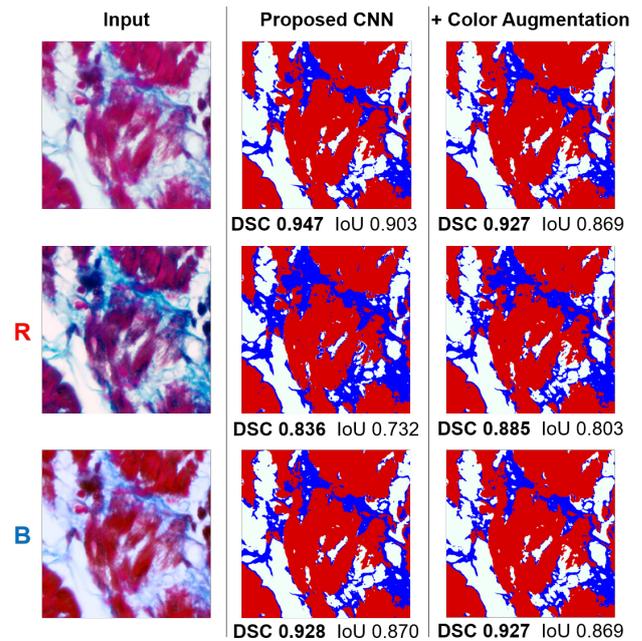

**Fig. 6** Comparison between predictions for the original image (*top*) and versions color-adjusted via the red (*middle*) and blue channels (*bottom*) demonstrates that the proposed method is robust to typical color variations in the input. *Left* to *right* columns: input image, predictions from the proposed CNN, and that with additional color-augmentation applied to training data. Mean Dice similarity coefficient (DSC) and mean intersection over union (IoU) are indicated for each corresponding image group

intensities, mean DSC of the proposed CNN decreased slightly. This test color modification did not impact the performance of the system trained on color-augmented images. However, performance of the proposed CNN without such training augmentation was better, by 0.001. Thus, training color augmentation did not enhance performance for this amount of change in test image color. The test images altered by changing red channel intensities exhibit strikingly larger visual dissimilarity to the original versions. For these images, the model trained on color-augmented images displayed markedly improved consistency in performance, though with the caveat of diminished accuracy on the original images. These results demonstrate robustness of the proposed system in coping with colors outside the range encountered during learning; that color augmentation during learning is not necessary if an evaluation dataset does not contain large disparities in color for certain structures, relative to training data.

### 3.2 Comparison with previous methods

Fig. 7 and Table 1 present a comparison between segmentation performance of the various methods. *K*-means clustering produced considerably inferior results compared to all CNNs we applied, scoring

**Table 1** Comparison between previous methods and the proposed CNN

| Method | Mean DSC | Mean Class DSC | | | Mean IoU | Mean Class IoU | | | Parameters |
|---|---|---|---|---|---|---|---|---|---|
| | | M | B | F | | M | B | F | |
| *k*-means RGB | 0.627 | 0.741 | 0.806 | 0.333 | 0.497 | 0.595 | 0.681 | 0.216 | - |
| *k*-means Lab | 0.575 | 0.717 | 0.768 | 0.241 | 0.449 | 0.564 | 0.629 | 0.153 | - |
| U-Net | 0.945 | 0.980 | 0.949 | 0.905 | 0.899 | 0.962 | 0.907 | 0.829 | ~31,000,000 |
| FCN-8 | 0.917 | 0.968 | 0.925 | 0.858 | 0.852 | 0.939 | 0.864 | 0.754 | ~134,000,000 |
| **Proposed CNN** | **0.947** | **0.980** | **0.955** | **0.907** | **0.903** | **0.961** | **0.916** | **0.832** | **~300,000** |

DSC – Dice similarity coefficient, IoU – intersection over union, M – myocyte, B – background, F – fibrosis.

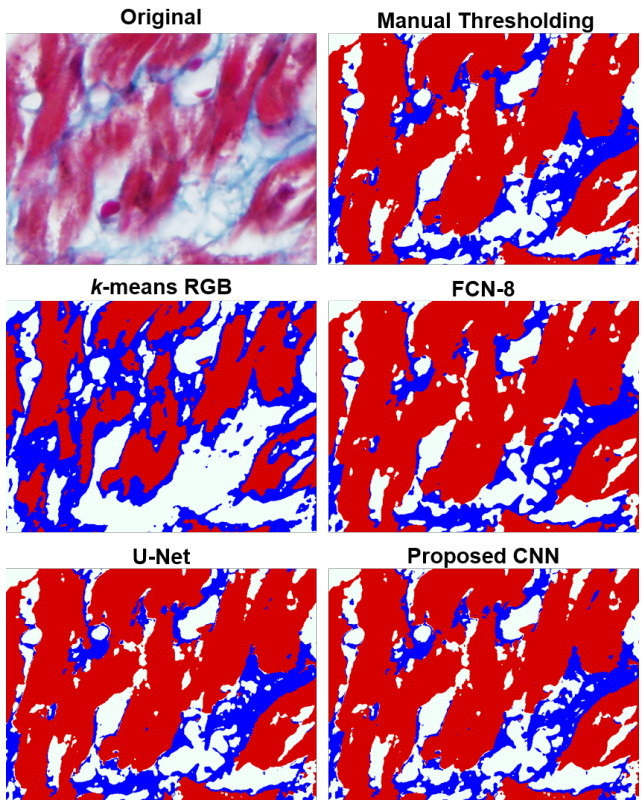

**Fig. 7** Representative examples of segmentations produced by manual thresholding (ground truth), *k*-means in RGB, FCN-8, U-Net, and the proposed method

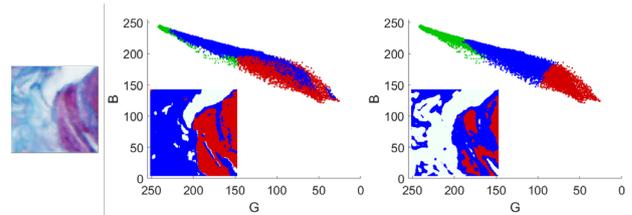

**Fig. 8** Illustration of the poor performance by *k*-means clustering for segmentation. Segmentations for the patch (*left*) by the proposed CNN (*center*) and *k*-means in RGB (*right*). Pixels are plotted according to their original RGB intensities. Background pixels are in green for clarity

poorly in all computed performance metrics. Predicted segmentation maps from *k*-means in RGB were characterized by fibrosis over-classification especially around edges of myocytes, and inconsistent fibrosis classification between unambiguously fibrotic regions of similar appearance. Performance scores across all metrics were lower when *k*-means was executed post conversion of images to Lab space, and predictions demonstrated similar types of inconsistencies (not shown).

To illuminate the shortfalls of this unsupervised learning algorithm, we visualize segmented class clusters according to original pixel RGB values in Fig. 8. The clusters segmented by the proposed CNN were highly irregular and interlaced in some regions. Contrastingly, clusters partitioned by *k*-means were well-separated and roughly equal in shape and volume.

Fig. 8 indicates that the images do not satisfy key assumptions of the *k*-means algorithm; namely spherical cluster distributions of similar variance, and equal class probabilities for every data point [56]. As well as lacking recognition of contextual and spatial information (unlike CNNs), *k*-means utilizes the same rigid Euclidean distance minimization for all classes, albeit inequality of cluster shape and size, or ill-defined boundaries. These limitations might be overcome by special transformations serving as preprocessing. However, determining suitable transformations is highly challenging. Since their success depends heavily on exact intensities, unique transformations tailored for different local regions are likely required, rendering this approach labor-intensive, and sensitive to inter- and intra-image intensity variations (similar to manual thresholding). Further, intensities are perceived by the human visual system in a highly nonlinear mapping, so although separation of the classes may be trivial by eye, the same task performed with only pixel values is much more difficult. In contrast to *k*-means, CNNs map highly complex relationships between intensities and desired outputs automatically, is robust to intensity variations (Fig. 6), and produce markedly superior results (Table 1 and Fig. 7). Despite being fully automated, these results strongly suggest that *k*-means clustering is unattractive for this application, primarily due to its inadequacy in segmenting similar images in their raw form to a satisfactory standard.

10Comparing segmentation performance of the three CNNs (Table 1), the proposed architecture is superior to both FCN-8 and U-Net. FCN-8 scored the lowest in performance out of the three CNNs by a relatively large margin. Segmentations by FCN-8 indicate a tendency to smooth structural edges and an inability of the network to capture fine details, often resulting in coarse and overly-large predicted areas for fibrosis (Fig. 7). Predictions from U-Net and the proposed network very closely resembled ground truth, indicating their capability to capture fine-grained details in the images precisely.

Furthermore, we assessed the relative efficiency of our proposed CNN against U-Net and FCN-8. Our architecture has 0.9 trillion FLOPs (multiply-adds) for a 2064 × 1536-pixel image with 3 channels. This is a 60% reduction to U-Net (2.3 trillion FLOPs), and a 30% reduction to FCN-8 (1.3 trillion FLOPs) for inputs of the same size (Fig. 9). Hence our model attains state-of-the-art performance with substantially less computation. It is worth noting that even though FCN-8 is more efficient than U-Net, its segmentation performance is clearly inferior (Fig. 7 and Table 1). Additionally, we observed stronger overfitting to training data by U-Net compared to the proposed model (which contains 103 times fewer parameters), as indicated by a much higher training accuracy relative to evaluation accuracy by the former.

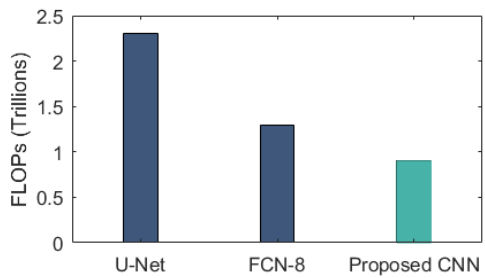

**Fig. 9** The proposed CNN performs competitively against the state-of-the-art with considerably greater efficiency, as measured by FLOPs (multiply-adds) for a 2064 × 1536-pixel input with 3 channels

Our proposed 11-layer CNN designed for per-pixel classification deviates from conventional forms particularly in its exclusion of subsampling layers (such as pooling) and upsampling layers, and its consistent number of intermediate feature channels through the network. We use only three types of functions (convolution, ReLU, and batch normalization) arranged in an uncomplicated configuration. The network's state-of-the-art results (Table 1) indicates that subsampling is not essential for data characterized by high variance of features between adjacent pixels and between different local regions, such as cardiac histological images. The aforementioned distinctions of the proposed CNN grant it several advantages over state-of-the-art architectures designed for segmentation: reduces the total number of learnable parameters, prevents overfitting to training data, improves efficiency, maintains image resolution, and allows fine-grained details to be captured accurately.

Design of a suitable neural network for an application can be difficult, as it involves selections from a vast number of possibilities, and there is no set of universal guidelines. Further, assessment of architectures and hyperparameters can be tedious. Many recent networks are characterized by substantial depth, or a diverse range of arrangements and connections between intermediate layers. Our work shows that shallower and uncomplicated networks can offer impressive performance. It may be worthwhile to start simple.

### 3.3 Atrial fibrosis under diabetes mellitus

Tallying the total pixels for each class in segmentation maps predicted by the proposed method and comparing control to DM images, each group consisted of 9.7% vs. 29.8% fibrosis, 63.2% vs. 44.5% myocytes, and 27.1% vs. 25.7% background. Thus, fibrosis greatly increased under DM by about a factor of 3, similar to the finding in [31]. Predicted quantifications were almost identical to those from the standard clinical approach; for example, manual thresholding predicted fibrosis of 10.1% and 28.7% in control and DM sets, respectively. Hence, the proposed method is validated for this application and may be exploited in a clinical context.

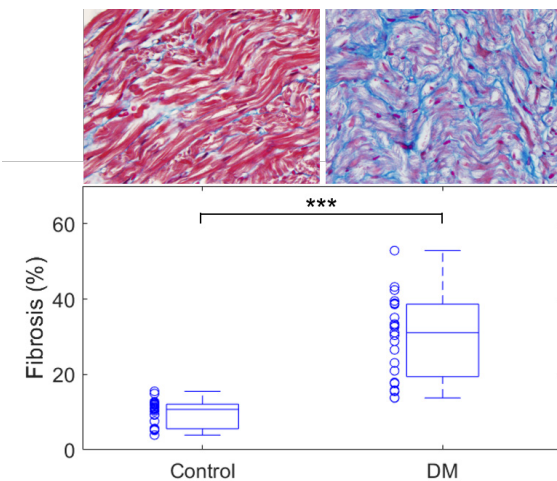

**Fig. 10** Comparison of per-image fibrosis percent area between control (*left*) and diabetes mellitus (*right*) rabbit atrial images ($n = 24$ each), as predicted by the 11-layer CNN. '***' = $p < 0.001$, two-sample $t$-test without assumption of equal variance. A representative image from each group is included

The difference between mean fibrosis proportions in control and DM images was unsurprisingly statistically significant (Fig. 10). Owing to the highly similar area of background in both groups, the additional fibrotic tissue was offset by approximately the same amount of decrease in myocytes, indicating direct replacement of cardiomyocyte regions by fibrosis. Since the DM rabbits (from which the images originated) were roughly 8-times more vulnerable to atrial fibrillation compared with controls [31], our results are in line with the proposed theories that structural remodeling under DM in the form of fibrosis may lead to electrophysiological remodeling and subsequent arrhythmia [33, 34].

## 4 Conclusion

In this paper, we proposed a novel 11-layer CNN and demonstrated the supervised learning-based approach for the application of histological image segmentation, particularly for fibrosis identification via Masson's trichrome staining. With an elegant configuration, drastically fewer parameters, and superior efficiency, the CNN outperformed the state-of-the-art on our image set. This approach is also robust to typical variations in image illumination and stain color. For best results, learning data should capture a rough representation of the characteristics in the total image set, including proportions of each class, and variations in color of certain structures.

**Acknowledgements** This work was supported by the National Heart Foundation of New Zealand and the Health Research Council of New Zealand (J.Z.).

## References

1. Csepe TA, Zhao J, Hansen BJ et al (2016) Human sinoatrial node structure: 3D microanatomy of sinoatrial conduction pathways. Prog Biophys Mol Biol 120(1-3):164-178
2. Hansen BJ, Zhao J, Fedorov VV (2017) Fibrosis and atrial fibrillation: Computerized and optical mapping; A view into the human atria at submillimeter resolution. JACC Clin Electrophysiol 3(6):531-546
3. Zhao J, Butters TD, Zhang H et al (2013) Image-based model of atrial anatomy and electrical activation: A computational platform for investigating atrial arrhythmia. IEEE Trans Med Imaging 32(1):18-27
4. Zhao J, Butters TD, Zhang H et al (2012) An image-based model of atrial muscular architecture: effects of structural anisotropy on electrical activation. Circ Arrhythm Electrophysiol 5(2):361-370
5. Gurcan MN, Boucheron LE, Can A, Madabhushi A, Rajpoot NM, Yener B (2009) Histopathological image analysis: A review. IEEE Rev Biomed Eng 2:147-171
6. Pham DL, Xu C, Prince JL (2000) Current methods in medical image segmentation. Annu Rev Biomed Eng 2(1): 315-337
7. Krizhevsky A, Sutskever I, Hinton GE (2012) ImageNet classification with deep convolutional neural networks. In: Pereira F, Burges CJC, Bottou L, Weinberger KQ (eds) Advances in neural information processing systems 25. Curran Associates Inc, Red Hook, pp 1097–1105
8. Ronneberger O, Fischer P, Brox T (2015) U-Net: convolutional networks for biomedical image segmentation. In: Navab N, Hornegger J, Wells W, Frangi A (eds) Medical image computing and computer-assisted intervention – MICCAI 2015. Lecture notes in computer science, vol 9351. Springer, Cham, p 234-241
9. Ciresan DC, Giusti A, Gambardella LM, Schmidhuber J (2013) Mitosis detection in breast cancer histology images with deep neural networks. In: Mori K, Sakuma I, Sato Y, Barillot C, Navab N (eds) Medical image computing and computer-assisted intervention – MICCAI 2013. Lecture notes in computer science, vol 8150. Springer, Berlin, Heidelberg, pp 411-418
10. Chen JM, Qu AP, Wang LW et al (2015) New breast cancer prognostic factors identified by computer-aided image analysis of HE stained histopathology images. Sci Rep 5:10690
11. Sertel O, Kong J, Catalyurek U, Lozanski G, Shanaah A, Saltz J, Gurcan MN (2008) Texture classification using nonlinear color quantization: Application to histopathological image analysis. Paper presented at the International Conference on Acoustics, Speech, and Signal Processing, Las Vegas, NV, 2008
12. Sieren JC, Weydert J, Bell A, De Young B, Smith AR, Thiesse J, Namati E, McLennan G (2010) An automated segmentation approach for highlighting the histological complexity of human lung cancer. Ann Biomed Eng 38(12):3581-3591
13. Wu G, Zhao X, Luo S, Shi H (2015) Histological image segmentation using fast mean shift clustering method. Biomed Eng Online 14(1):14-24
14. Meas-Yedid V, Tilie S, Olivo-Marin J-C (2002) Color image segmentation based on Markov random field clustering for histological image analysis. Paper presented at the 16th International Conference on Pattern Recognition, Quebec City, Quebec, 11-15 Aug 2002
15. Fukushima K (1980) Neocognitron: A self-organizing neural network model for a mechanism of pattern recognition unaffected by shift in position. Biol Cybern 36(4):193-202
16. Hubel DH, Wiesel TN (1968) Receptive fields and functional architecture of monkey striate cortex. J Physiol 195(1):215-243
17. Nair V, Hinton GE (2010) Rectified linear units improve restricted Boltzmann machines. In: ICML'10 Proceedings of the 27th International Conference on International Conference on Machine Learning, pp 807-814
18. Simonyan K, Zisserman A (2015) Very deep convolutional networks for large-scale image recognition. Paper presented at the International Conference on Learning Representations, San Diego, CA, 7-9 May 2015
19. He K, Zhang X, Ren S, Sun J (2015) Deep residual learning for image recognition. Paper presented at the IEEE Conference on Computer Vision and Pattern Recognition (CVPR), Las Vegas, NV, 27-30 June 2016
20. Poznanski A, Wolf L (2016) CNN-N-Gram for Handwriting Word Recognition. Paper presented at the IEEE Conference on Computer Vision and Pattern Recognition (CVPR), Las Vegas, NV, 27-30 June 2016
21. Ren S, He K, Girshick R, Sun J (2017) Faster R-CNN: Towards real-time object detection with region proposal networks. IEEE Trans Pattern Anal Mach Intell 39(6):1137-1149
22. Farabet C, Couprie C, Najman L, Lecun Y (2013) Learning hierarchical features for scene labeling. IEEE Trans Pattern Anal Mach Intell 35(8):1915-1929
23. Badrinarayanan V, Kendall A, Cipolla R (2017) SegNet: A deep convolutional encoder-decoder architecture for image segmentation. IEEE Trans Pattern Anal Mach Intell 39(12):2481-2495
24. Liskowski P, Krawiec K (2016) Segmenting retinal blood vessels with deep neural networks. IEEE Trans Med Imaging 35(11):2369-2380
25. Pereira S, Pinto A, Alves V, Silva CA (2016) Brain tumor segmentation using convolutional neural networks in MRI images. IEEE Trans Med Imaging 35(5):1240-1251
26. Xu J, Luo X, Wang G, Gilmore H, Madabhushi A (2016) A deep convolutional neural network for segmenting and classifying

26. epithelial and stromal regions in histopathological images,. Neurocomputing 191:214-223
27. Cruz-Roa, A, Basavanhally A, González F et al (2014) Automatic detection of invasive ductal carcinoma in whole slide images with convolutional neural networks. In: Proceedings of SPIE, 2014
28. Zeisberg EM, Potenta SE, Sugimoto H, Zeisberg M, Kalluri R (2008) Fibroblasts in kidney fibrosis emerge via endothelial-to-mesenchymal transition. J Am Soc Nephrol 19(12):2282-2287
29. Marks LB, Yu X, Vujaskovic Z, Small W Jr, Folz R, Anscher MS (2003) Radiation-induced lung injury. Semin Radiat Oncol 13(3):333-345
30. Chang TT, Liaw YF, Wu SS et al (2010) Long-term entecavir therapy results in the reversal of fibrosis/cirrhosis and continued histological improvement in patients with chronic hepatitis B. Hepatology 52(3):886-893
31. Fu H, Li G, Liu C, Li J, Wang X, Cheng L, Liu T (2015) Probucol prevents atrial remodeling by inhibiting oxidative stress and TNF-α/NF-κB/TGF-β signal transduction pathway in alloxan-induced diabetic rabbits. J Cardiovasc Electrophysiol 26(2):211-222
32. Kannel WB, Wolf PA, Benjamin EJ, Levy D (1998) Prevalence, incidence, prognosis, and predisposing conditions for atrial fibrillation: Population-based estimates. Am J Cardiol 82(7):2N-9N
33. Fu H, Li G, Liu C et al (2016) Probucol prevents atrial ion channel remodeling in an alloxan-induced diabetes rabbit model. Oncotarget 7(51):83850–83858
34. Kato T, Yamashita T, Sekiguchi A et al (2008) AGEs-RAGE system mediates atrial structural remodeling in the diabetic rat. J Cardiovasc Electrophysiol 19(4):415-420
35. Nattel S, Burstein B, Dobrev D (2008) Atrial remodeling and atrial fibrillation: Mechanisms and implications. Circ Arrhythm Electrophysiol 1(1):62-73
36. Zhao J, Hansen BJ, Wang Y et al (2017) Three-dimensional integrated functional, structural, and computational mapping to define the structural 'fingerprints' of heart-specific atrial fibrillation drivers in human heart ex vivo. J Am Heart Assoc 6(8)
37. Long J, Shelhamer E, Darrell T (2017) Fully convolutional networks for semantic segmentation. IEEE Trans Pattern Anal Mach Intell 39(4):640-651
38. Noh H, Hong S, Han B (2015) Learning deconvolution network for semantic segmentation. Paper presented at the IEEE Conference on Computer Vision (ICCV), Santiago, Chile, 7-13 Dec 2015
39. Badrinarayanan V, Handa A, Cipolla R (2015) SegNet: A deep convolutional encoder-decoder architecture for robust semantic pixel-wise labelling. arXiv preprint arXiv:1505.07293
40. Zeiler MD, Fergus R (2013) Visualizing and understanding convolutional networks. arXiv preprint arXiv:1311.2901
41. Ioffe S, Szegedy C (2015) Batch normalization: Accelerating deep network training by reducing internal covariate shift. In: ICML'15 Proceedings of the 32nd International Conference on International Conference on Machine Learning, pp 448-456
42. Srivastava N, Hinton G, Krizhevsky A, Sutskever I, Salakhutdinov R (2014) Dropout: A simple way to prevent neural networks from overfitting. J Mach Learn Res 15:1929-1958
43. Yu F, Koltun V (2016) Multi-scale context aggregation by dilated convolutions. Paper presented at the International Conference on Learning Representations (ICLR), 2016
44. Howard AG (2013) Some improvements on deep convolutional neural network based image classification. arXiv preprint arXiv:1312.5402
45. Japkowicz N, Stephen S (2002) The class imbalance problem: A systematic study. Intell Data Anal 6(5):429-449
46. Weiss G, Provost F (2001) The effect of class distribution on classifier learning: an empirical study. Technical report, Rutgers University
47. TensorFlow (2018). https://www.tensorflow.org. Accessed 14 Mar 2018
48. He K, Zhang X, Ren S, Sun J (2015) Delving deep into rectifiers: Surpassing human-level performance on ImageNet classification. Paper presented at the International Conference on Computer Vision (ICCV), Santiago, Chile, 7-13 Dec 2015
49. Kingma DP, Ba JL (2015) Adam: A method for stochastic optimization. arXiv preprint arXiv:1412.6980v9
50. Zou KH, Warfield SK, Bharatha A et al (2004) Statistical validation of image segmentation quality based on a spatial overlap index. Acad Radiol 11(2):178-189
51. Everingham M, Eslami SM, Gool L, Williams CK, Winn J, Zisserman A (2015) The PASCAL visual object classes challenge: A retrospective. Int J Comput Vis 111(1):98-136
52. Sirinukunwattana K, Pluim JPW, Chen H et al (2017) Gland segmentation in colon histology images: The GlaS challenge contest. Med Image Anal 35:489-502
53. Salehi SSM, Erdogmus D, Gholipour A (2017) Auto-context convolutional neural network (Auto-Net) for brain extraction in magnetic resonance imaging. IEEE Trans Med Imaging 35(11):2319-2330
54. Jansson A, Humphrey E, Montecchio N, Bittner R, Kumar A, Weyde T (2017) Singing voice separation with deep U-Net convolutional networks. Paper presented at the 18th International Society for Music Information Retrieval Conference, Suzhou, China, 2017
55. MacQueen J (1967) Some methods for classification and analysis of multivariate observations. In: Proceedings of the Fifth Berkeley Symposium on Mathematical Statistics and Probability. University of California Press, Berkeley, pp 281-297
56. Jain AK (2010) Data clustering: 50 years beyond K-means. Pattern Recognit Lett 31(8):651-666

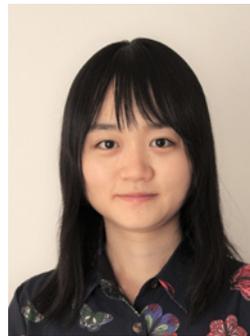

**Xiaohang Fu** received her Bachelor of Engineering (specializing in Biomedical Engineering) with First Class Honors in 2018 from The University of Auckland, New Zealand. She was a research assistant at the Auckland Bioengineering Institute. Her research interests include application of machine learning to computer vision problems, and computational neuroscience.

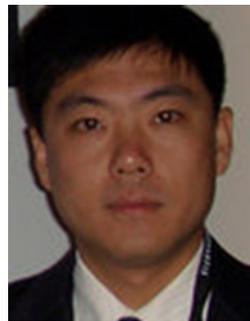

**Dr. Tong Liu, PhD, MD**, is an Associate Professor of Cardiology and Deputy Director of the atrial fibrillation centre at Tianjin Institute of Cardiology in Tianjin Medical University, and an electrophysiologist in the Department of Cardiology, Hospital of Tianjin Medical University. As an electrophysiologist, his principal duties are focused on cardiac electrophysiology, especially in the field of AF. His research interests concern the study of AF and arrhythmias extending from basic sciences to clinical medicine, especially his international leading research using small animal diabetic models to understand the basic mechanisms of inflammation and oxidative stress in the pathogenesis of atrial fibrillation under diabetic conditions and associated upstream therapies for AF prevention using conventional medicines.

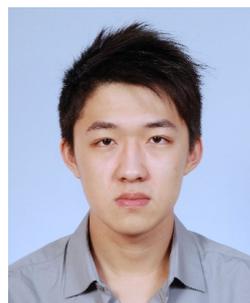

**Zhaohan Xiong** is a research assistant at Auckland Bioengineering Institute, The University of Auckland. He is also in his final year of his Honors degree for a Bachelor of Engineering, specializing in optimization and analytics. His research focus is to apply data science techniques to solve problems involving medical imaging and signal processing.

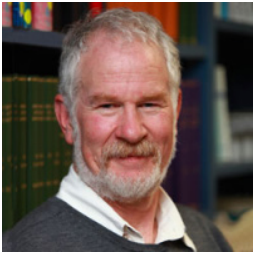
**Bruce Smaill** is a Principal Investigator in the Auckland Bioengineering Institute, University of Auckland, New Zealand, and a long-standing member of the Department of Physiology. He is interested in the muscular architecture of the heart and how this affects electrical and mechanical function in normal and diseased hearts. His research with colleagues in Auckland has been influential in the field. It combines structural imaging, experimental studies and computer modelling, and involves a team of physiologists, bioengineers and clinicians.

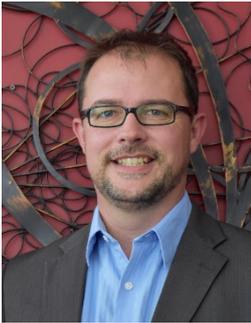
**Martin Stiles** trained at the University of Otago and returned to Hamilton after a year working at Dunedin Hospital. His initial training in Cardiology at Waikato Hospital was completed at the Royal Infirmary of Edinburgh. Training in Electrophysiology was continued at Glenfield Hospital, Leicester and at Green Lane Cardiovascular Service. In 2005 he was awarded a New Zealand Heart Foundation Overseas Training Fellowship to commence a PhD on Atrial Fibrillation and Flutter at the University of Adelaide, before returning to Hamilton in 2008. At Waikato Hospital and Midland Cardiovascular Services (located at Braemar Hospital) he established a programme of Atrial Fibrillation Ablation and, with the purchase of a 3D mapping system, advanced the service to include further complex radiofrequency ablation including treatment of Ventricular Tachycardia. Since 2012, cryoablation is also offered at these centres. Martin maintains an active interest in cardiac devices and he implants Pacemakers, Defibrillators and Resynchronisation Devices.

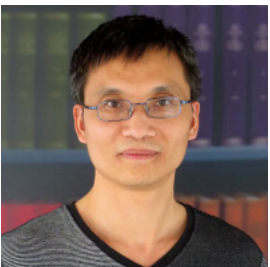
**Jichao Zhao** received his Bachelor and Masters of Science in Computational Mathematics from Northeastern University, China. In 2006, Jichao obtained a PhD in Applied Mathematics at the University of Western Ontario, Canada. Jichao then worked one year in INRIA, France as a postdoctoral fellow. In September of 2007, Jichao joined the Auckland Bioengineering Institute at The University of Auckland, New Zealand, and is Senior Research Fellow. His current research interests center on development and application of novel computational approaches including image-based computer models, signal processing and deep learning to study the basic mechanisms underlying atrial fibrillation.